\documentclass{article}
\usepackage{spconf,amsmath,graphicx}
\usepackage{flushend}
\usepackage[norelsize, linesnumbered, ruled, lined, boxed, commentsnumbered]{algorithm2e}
\usepackage{hyperref}

\title{Audio visual character profiles for detecting background characters in entertainment media}
\name{Rahul Sharma, Shrikanth Narayanan}
\address{University of Southern California, USA}
\begin{document}
%
\maketitle
\begin{abstract}
An essential goal of computational media intelligence is to support understanding how media stories -- be it news, commercial or entertainment media -- represent and reflect society and these portrayals are perceived. People are a central element of  media stories.  This paper focuses on understanding the representation and depiction of background characters in media depictions, primarily movies and TV shows. We define the background characters as those who do not participate vocally in any scene throughout the movie and address the problem of localizing background characters in videos. We use an active speaker localization system to extract high-confidence face-speech associations and generate audio-visual profiles for talking characters in a movie by automatically clustering them. Using a face verification system, we then prune all the face-tracks which match any of the generated character profiles and obtain the background character face-tracks. We curate a background character dataset  which provides annotations for background character for a set of TV shows, and use it to evaluate the performance of the background character detection framework.
\end{abstract}
\section{Introduction}
\label{sec:intro}

Technological advances have made it easier than ever to produce and experience media content, across platforms and genres, be it news, movies, TV shows, digital shorts or user generated videos. 
Media consumption has become an integral part of our lives in this modern world. This calls for a pressing need to develop methods to analyze and understand the impact of media content on human life, be it societal or economic. There has been a recent surge of research under the theme of computational media intelligence \emph{CMI}, which focus on providing tools to analyze the people, places, and context in the multimedia content to gain a holistic understanding of the stories being told, and their impact on the experiences and behavior of individuals and society at large \cite{SomComputationalMediaIntelligence:Human-centered}. 

One of the goals of CMI relates to the representation and identity of the people (\emph{characters}) depicted in the media, and understanding their portrayals in the stories. Characters in media stories can be broadly divided into i) primary or lead characters, ii)  background characters who appear in the background of a scene and participate minimally in an ongoing interaction. There have been numerous works dealing with detecting and localizing instances of primary characters in entertainment media, primarily TV shows and movies\cite{bojanowski2013finding,haurilet2016naming}. In contrast, learning representations for background characters and studying their portrayals is relatively unexplored~\cite{cour2011learning, parkhi2018automated, bojanowski2013finding}. In this work, we address the problem of detecting background characters in media content, particularly movies and TV shows. 
In entertainment media, primary characters appear most of the time and speak very often. Background characters on the other hand may appear for a small fraction of time, and may not speak during the entire movie. In this work, we operationally define background characters as those who do not speak throughout the movie. 

This paper presents an unsupervised framework to detect background character face tracks in videos, enlisting the face tracks that do not show speech activity throughout the video.
Since there has been little past research on computationally studying the characteristics of background characters, there are no readily available large-scale datasets for this purpose. 
Hence in this work we leverage an existing dataset \cite{VPCD} consisting of primary characters that have been annotated for some selected TV shows and movies to derive a complementary set of background character face tracks for evaluation. For validating  the automatically obtained set of background character face tracks, we post processed this dataset by verifying using  manual annotations (to create the final evaluation set). 
The proposed framework's performance is evaluated on this newly curated background character dataset.

We build on recent advances in active speaker localization~\cite{sharma2020cross, ava, icip_rahul} in unconstrained videos and audio speaker verification~\cite{chung2018voxceleb2, wan2018generalized} strategies to develop audio-visual character profiles for the talking characters involved in a movie. Rather than collecting all the instances of characters' speaking, we construct generic audio-visual representations for the characters in a movie. Further using face verification techniques~\cite{schroff2015facenet,retinaface} we collect all the face tracks that do not match the created audio-visual character profiles and call them background characters. 

The main contributions of this paper include developing  i) method assisting the visual signal-based active speaker localization system with speaker recognition information,  and show enhanced performance ii)  an iterative clustering paradigm to construct audio-visual character profiles for primary characters in movies iii) a background character dataset, which can be used to understand the representation and characteristics of background characters in media content.

\section{Background Character Dataset}
\label{sec:bchar_dataset}

\begin{figure}
    \centering
    \includegraphics[width=0.45\textwidth,keepaspectratio]{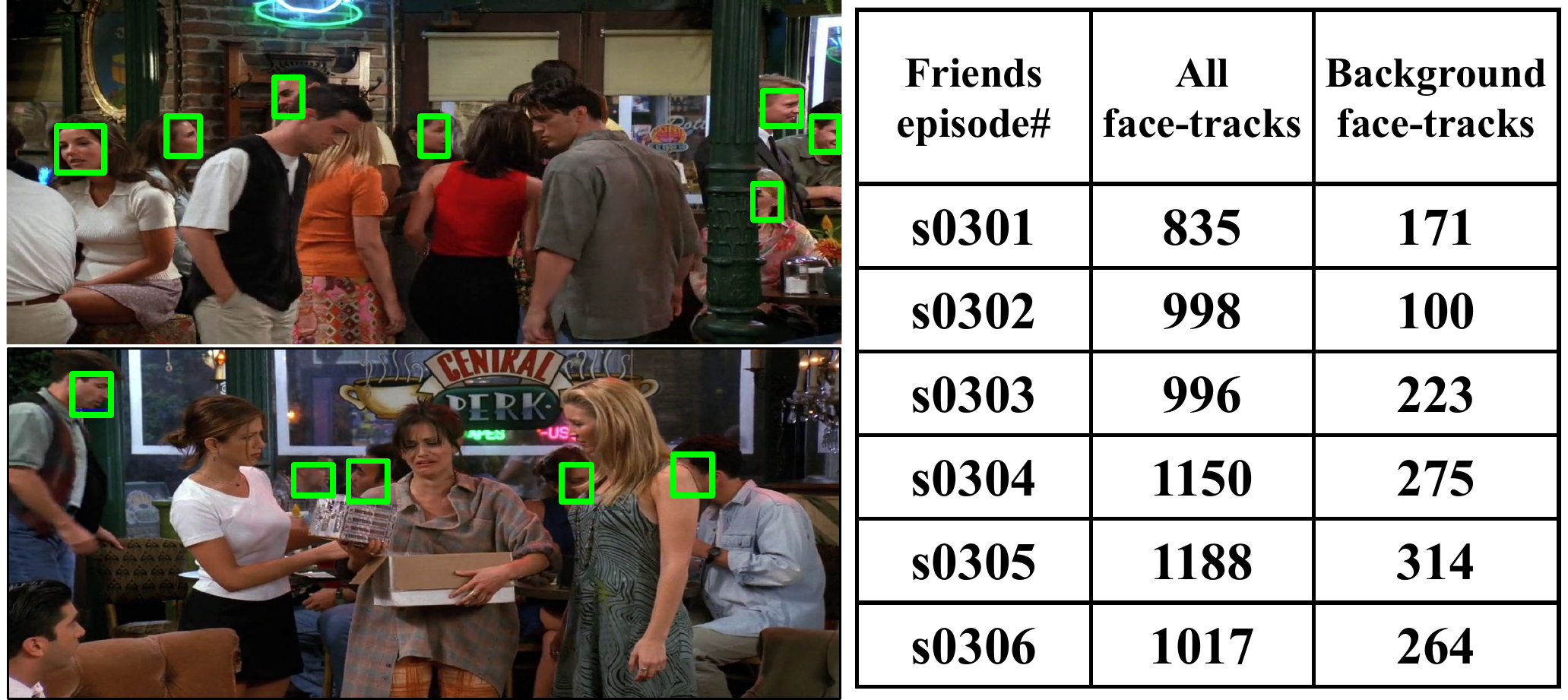}
    \caption{\emph{Left:} Sample frames with background characters marked in green. \emph{Right:} Number of tracks in each episode.}
    \label{fig:bchar}
\end{figure}

This paper introduces a pilot dataset consisting of background characters' face tracks for episodes from the sitcom {\em Friends}. We targeted to obtain the background character face-tracks as a complementary set of primary character face-tracks. We use the Visual Person Clustering Dataset~\cite{VPCD} (VPCD), which provides annotations for all the characters involved in conversations (speaking at some point in time) and the timestamps of their corresponding voice activity along with the character IDs. We use RetinaFace~\cite{retinaface} and MM track~\cite{mmtrack2020} to get the face-tracks in the video. We then purge all the primary character face-tracks (obtained from VPCD), which overlap with the obtained face-tracks in space and time. To further refine the dataset, we remove the face tracks which matches a primary character with high confidence, using VGGFace-2~\cite{vggface2} to obtain face-track embeddings. As a final refinement step, we obtain manual annotations. Figure~\ref{fig:bchar} shows details about the curated dataset.

\section{Proposed Approach}

\subsection{Problem formulation}
Given a speaker homogeneous voiced segment $v$ and a corresponding set of temporally overlapping face-tracks $\{f_k\}$ ($k \in [1,K]$), we aim to find a score for the mapping $v \rightarrow f_k$, signifying face-track $f_k$ is the source of the speech segment $v$. We obtain the active voice regions in a movie, using a speech segmentation tool pyannote~\cite{pyannote}, and partition them at the shot boundaries, which we gather using PyScenedetect. The partitioning at the shot boundaries is motivated by the fact that speaker change is a prominent movie cut attribute~\cite{moviecuts}, thus decreasing the likelihood of observing a speaker change in the obtained segments. We further partition the voice segments to have a maximum duration of $1s$, which further increases the probability of the acquired voiced segments $v_n$ ($n \in [1,N]$) being speaker homogeneous.

We use RetainFace~\cite{retinaface} to obtain the set of face-tracks $\{f_{k}\}$ from visual frames that coincide with the speech segment $v_n$ in time. Now, for each speech segment $v_n$, we have $K$ potential active speaker instances, $\{v_n, f_k\}$, where $k \in [1, K]$ and $n \in [1,N]$, signifying face-track $f_k$ is a potential source for speech segment $v_n$. We describe the methodology to compute a score for each instance using the audio-visual features in the following sections. 

\subsection{Visual active speaker score}
In a recent work, we introduced Hierarchichally context-aware~\cite{sharma2020cross} (HiCA) cross-modal network, trained to detect the presence of speech in a video segment. It was demonstrated, using class activation maps (CAMs)~\cite{zhou2015cnnlocalization} for the positive class, that such a system has the ability to localize active speaker faces in a video reliably.
We use the HiCA network, pretrained on movies, to obtain CAMs for a video. We further compute visual active speaker (VAS) score for all face-tracks as the aggregated mean of the CAMs pertaining to the region of interest, as denoted below. $\text{CAM}_f$ represents the CAMs for frame $f$ and $Z$ denotes the averaging factor.
\begin{equation}
\label{eq:cam}
    \text{VAS}_{k} = \frac{1}{Z}\Sigma_{f}\Sigma_{x,y}(\text{CAM}_f[y_1:y_2, x_1:x_2])
\end{equation}
We select the speech segments with only one face-track associated with the visual stream with a high VAS score and call them high-confidence instances (HCI). Essentially,
\begin{equation}
    \text{HCI} \equiv \{v_n, f_k\} \forall n\in[1,N]: \text{VAS}_k > \tau, K= 1
\end{equation}

\subsection{Profile matching score}
We intend to generate individual speech and face profiles for speaking characters. We impose the constraint that given an active speaker instance $\{v_n, f_k\}$, if $f_k$ matches one character profile, the speech segment, $v_n$ should align with the corresponding speech profile of the character.
We represent a face-track, $f_k$, using average of the involved face embeddings, obtained using a ResNet50 trained on VGGFace-2~\cite{vggface2}. To generate the speech segment representation, we used the speaker recognition model by pyannote~\cite{pyannote}.

To generate the character speech-face profiles, we cluster the instances in HCI, using the visual modality. We use the Hierarchical-DBSCAN~\cite{mcinnes2017hdbscan} for clustering, as it requires no pre-specified cluster number and offers a way to compute soft cluster participation at inference time. We cluster the instances in HCI, to obtain $L$ clusters ${F_l}$, where $l \in [1, L]$. We refer to the set of corresponding speaker embeddings for points in $F_l$ as $V_l$.
We call $F_l$ and $V_l$ the face and speech profiles, respectively.

For a speech-face instance, $\{v_n, f_k\}$ we compute the profile matching score (PMS) as $P(v_n \rightarrow f_k)$, f $f_k$ being the sound source for speech segment $v_n$, as:
\begin{align}
    P(v_n \rightarrow f_k) = \Sigma_lP(v_n \in V_l, f_k \in F_l) \\
    P(v_n \rightarrow f_k) = \Sigma_lP(v_n \in V_l)P(f_k \in F_l)
\end{align}
which follows from the fact that $v_n$ and $f_k$ independently share the membership with cluster $l \in [1,L]$ in corresponding modalities.

We estimate $P(f_k \in F_l)$, using the outlier-based membership score from hDBSCAN, which uses the GLOSH~\cite{glosh} algorithm to compute the outlier score. Outlier-based membership score facilitates inference in the case of unseen characters. To estimate the membership for $v_n$, we first cluster the $v_n \forall n \in \text{HCI}$ using the hDBSCAN to obtain $B$ clusters $\Tilde{V}_b$, where $b \in [1,B]$. Using outlier-based membership score for $P(v_n \in \Tilde{V_b})$ from hDBSCAN:
\begin{align}
    P(v_n \in V_l) = \Sigma_bP(v_n \in V_l | v_n \in \Tilde{V_b})P(v_n \in \Tilde{V_b}) \\
    P(v_n \in V_l | v_n \in \Tilde{V_b}) = \frac{|V_l \cap  \Tilde{V_b}|}{|\Tilde{V_b}|} 
\end{align}
which follows from the total probability rule and $|.|$ denotes the cardinality of the set. Thus we estimate PMS $=$
\begin{align}
\label{eq:PMS}
     \Sigma_l\Sigma_bP(v_n \in V_l | v_n \in \Tilde{V_b})P(v_n \in \Tilde{V_b})P(f_k \in F_l)
\end{align}
\begin{algorithm}[tb]
\label{iterative  profile matching}
$\text{HCI} \gets \{s_n, f_k\} \forall n\in[1,N]: \text{VAS}_k > \tau, K= 1$\;
$iter = 0$\;
\While{$|\text{HCI}|$ increases}{
    $iter = iter + 1$\;
    $(F_l, V_l) \gets \text{HCI}$ \tcp*{Clustering}
    \For{each instance $\{s_n, f_k\} \not\in \text{HCI}$}{
    Compute $\text{VAS}_k$ \tcp*{using eq.~\ref{eq:cam}}
    Compute $\text{PMS}$ \tcp*{using eq.~\ref{eq:PMS}}
    $\alpha = 1 - 0.95^{iter}$\;
    $score = \alpha\text{PMS} + (1-\alpha)\text{VAS}_k$\;
    \If{$\text{score} > \delta$ }
    {
        add instance ${s_n, f_k}$ to HCI\;
    }
    }
  }
  \caption{Iterative profile matching algorithm}
\end{algorithm}
\subsection{Iterative profile matching}
Using equation~\ref{eq:cam}, we start with an initial set of high confidence speech-face associations, HCI, derived from the visual signal. In each iteration, we compute a confidence score, a linear combination of VAS and PMS, for each instance $\{v_n, f_k\}$. We add the instances with a high enough confidence score to the set HCI and repeat the algorithm. The paradigm is shown in algorithm~\ref{iterative  profile matching}.

In the initial iterations of the algorithm, since the number of high confidence instances is smaller, it likely that some characters do not have any instances in HCI. Since the cluster membership is fundamental to PMS, having no instances in HCI gives unreliable PMS scores. To alleviate the issue, we decrease the weight of the PMS scores in initial iterations, as shown in algo\ref{iterative  profile matching}:line8.

\begin{table}[b]
\centering
\caption{Performance for active speaker localization using audio-visual character profiles.}
\label{tab:active_speaker}
\resizebox{0.38\textwidth}{!}{%
\begin{tabular}{ccc}
\hline
Video Name      & \begin{tabular}[c]{@{}c@{}}CAMs\\ (auROC)\end{tabular} & \begin{tabular}[c]{@{}c@{}}Audio -visual\\ (auROC)\end{tabular} \\ \hline
Friends\_s03e01 & 0.77                                                   & 0.87                                                          \\ \hline
Friends\_s03e02 & 0.75                                                   & 0.77                                                          \\ \hline
Friends\_s03e03 & 0.70                                                   & 0.77                                                          \\ \hline
Friends\_s03e04 & 0.74                                                   & 0.80                                                          \\ \hline
Friends\_s03e05 & 0.74                                                   & 0.86                                                          \\ \hline
Friends\_s03e06 & 0.79                                                   & 0.84                                                          \\ \hline
\end{tabular}%
}
\end{table}
\subsection{Background character detection}
\label{subsec:bchar_detect}
With the premise that all primary characters speak for considerable times in a movie, it is highly likely that after enough iterations of profile matching, all primary characters have significant instances in the set of high confidence instances (HCI). Post iterative matching algorithm, we obtain character profiles, ${F_l, V_l}$ for $l \in [1, L]$, which we represent by the mean of the instances they possess. For any face-track $f_k$, we classify it as a background character track if its minimum distance from any character profile is more significant than a threshold, $\beta$.

\section{Performance Evaluation}
\subsection{Active speaker localization}
We use 6 episodes of the 3rd season of the TV show {\em Friends} for evaluation. We obtain the active-speaker ground truth from VPCD corpus \cite{VPCD}, where they provide the voice activity details and face-tracks for all the characters. We used algorithm~\ref{iterative  profile matching}, to obtain a score (weighted sum of VAS and PMS) for face-tracks that coincide with the voiced segments. For the face-tracks that do not overlap with any active voice regions, we score them with the corresponding $\text{VAS}/10$, scaling down to account for no voice activity. We extend the same face-track score to all the involved face-boxes and compare the performance against the ground truth. We report the area under the ROC curve for the two extreme steps of the algorithm: i) the first iteration, using just the VAS score, and ii) the last iteration, tabulated in Table~\ref{tab:active_speaker}. We note a clear improvement in the active speaker performance with the character profile matching, which is consistent across all the episodes.

\begin{figure}
    \centering
    \includegraphics[width=0.37\textwidth,keepaspectratio]{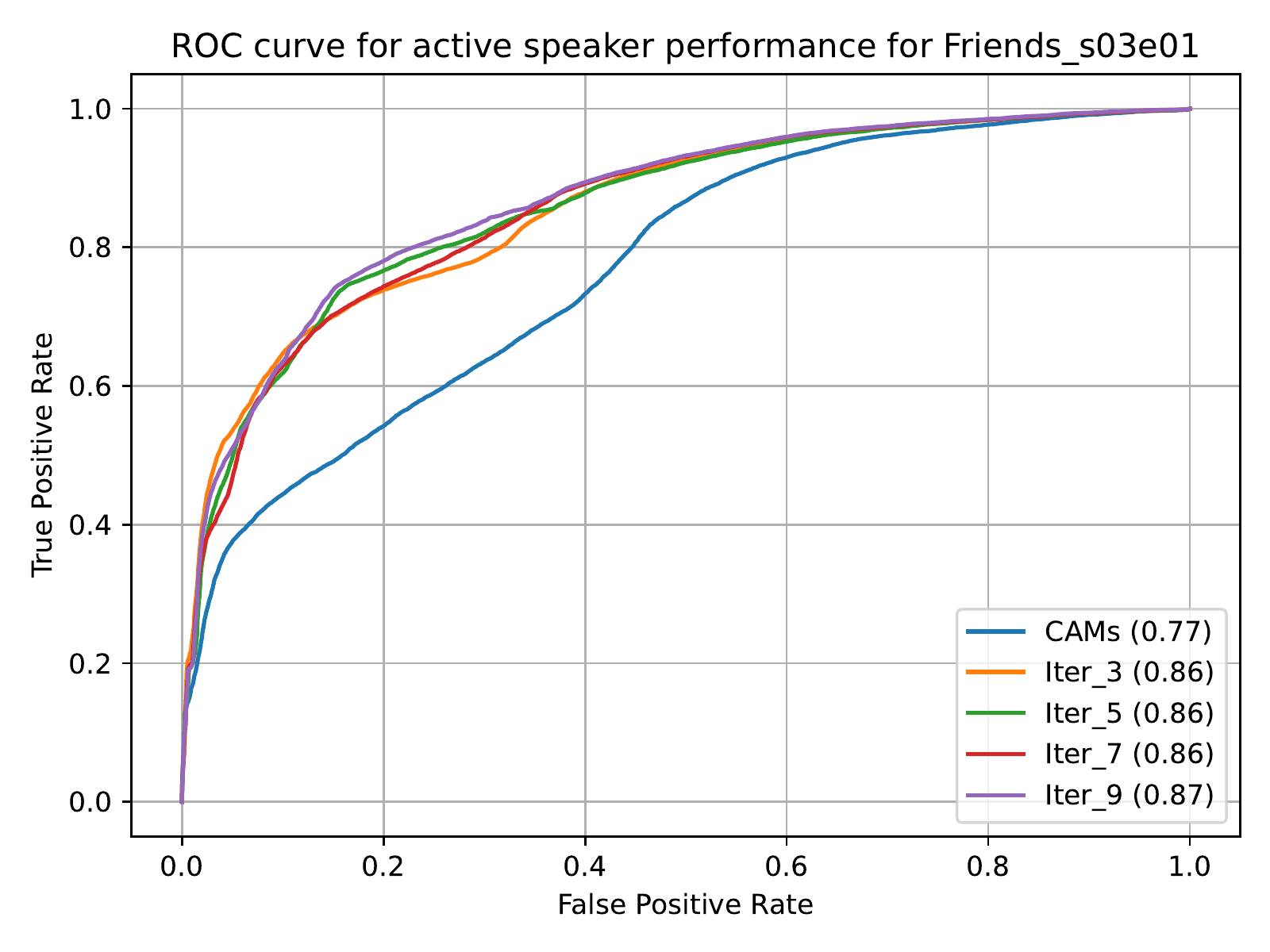}
    \caption{Performance for active speaker localization (ROC) for different iteration of profile matching algorithm.}
    \label{fig:roc}
\end{figure}

\begin{figure}
    \centering
    \includegraphics[width=0.37\textwidth,keepaspectratio]{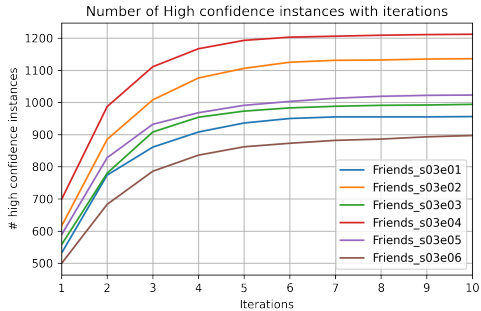}
    \caption{Increase in high confidence instances saturates with iterations.}
    \label{fig:num_HCI}
\end{figure}

\begin{figure}
    \centering
    \includegraphics[width=0.38\textwidth,keepaspectratio]{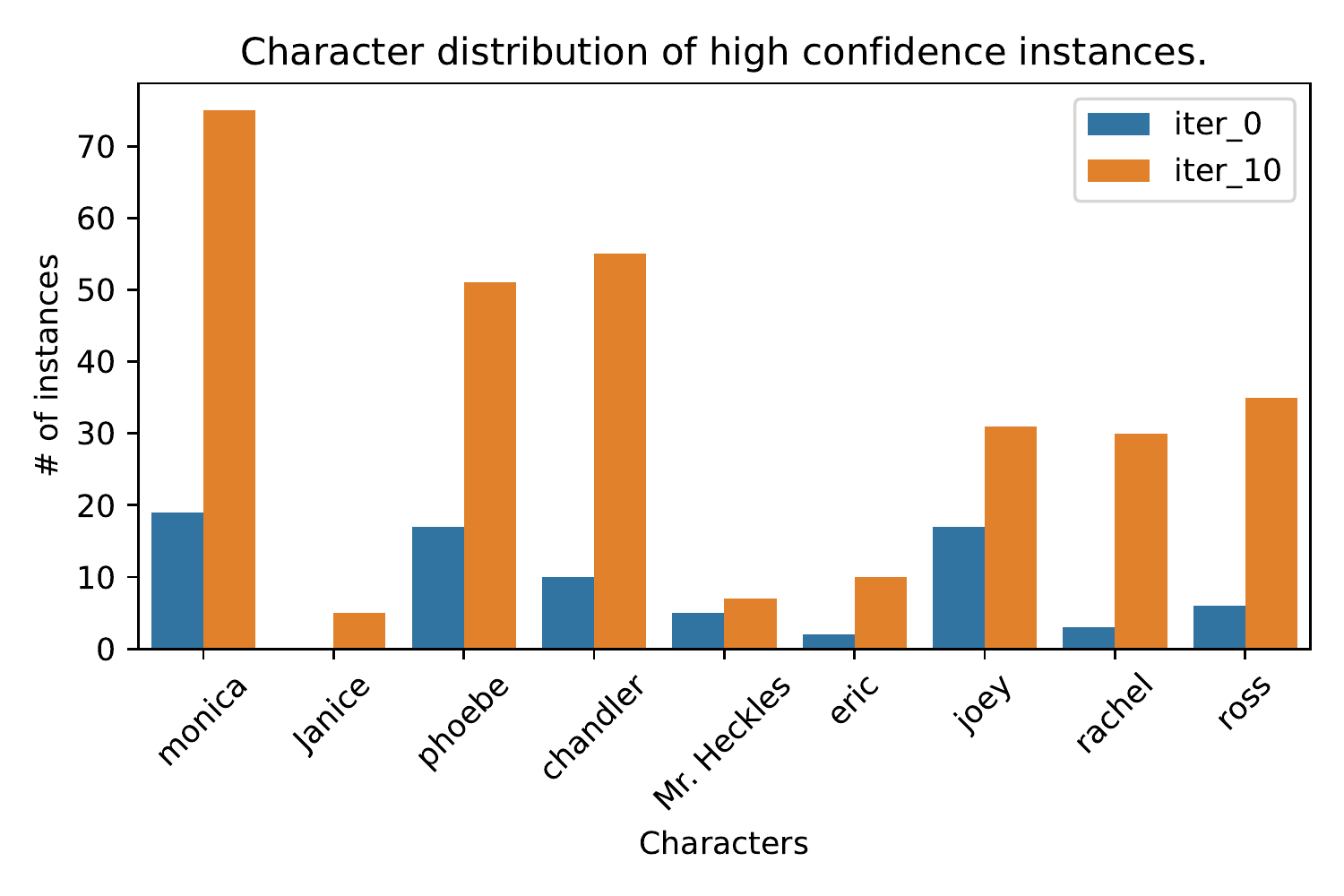}
    \caption{Distribution of characters among high confidence instances (HCI), for two extreme steps of iterative profile matching.}
    \label{fig:histogram}
\end{figure}

 To further understand the effect of the iterative algorithm, we study the performance of active speaker localization across different iterations. In figure~\ref{fig:roc} we show the performance, in terms of ROC, for \emph{Friends\_s03e01}, for every other iteration. We observed a significant jump in the initial iterations, which saturate as we further iterate. This can be attributed to the saturation in the number of high confidence instances with iterations, as shown in figure~\ref{fig:num_HCI}. We show the distribution of primary characters among the instances in HCI for the initial and terminating iterations, in fig~\ref{fig:histogram}. The notable increase in the instances across all the characters justifies the gained robustness in the character profiles, further supporting performance improvement. 

\subsection{Background character detection}
We use the newly curated background character dataset, as described in section~\ref{sec:bchar_dataset} for validation. As described in section~\ref{subsec:bchar_detect}, we obtain a score for each face-track in the video: the minimum distance from the acquired character profiles. In Table~\ref{tab:bchar} we show the performance of character profiles to detect background characters in terms of area under ROC. We further compute the ground truth character profiles by accumulating the face-track embeddings across the ground truth face-tracks provided by VPCD for all the primary characters. Using the computed ground-truth character profiles, we calculate the performance for detecting background characters as shown in Table~\ref{tab:bchar}. Against comparison to ground-truth character profiles, the estimated character profiles, using the iterative profile matching algorithm, performs well in detecting background characters.

\begin{table}[tb]
\centering
\caption{Performance for background character detection.}
\label{tab:bchar}
\resizebox{0.38\textwidth}{!}{%
\begin{tabular}{ccc}
\hline
Video Name & \begin{tabular}[c]{@{}c@{}}Character Profiles\\ auROC\end{tabular} & \begin{tabular}[c]{@{}c@{}}GT representation\\ auROC (max)\end{tabular} \\ \hline
Friends\_s03e01 & 0.82 & 0.88 \\ \hline
Friends\_s03e02 & 0.63 & 0.87 \\ \hline
Friends\_s03e03 & 0.79 & 0.90 \\ \hline
Friends\_s03e04 & 0.78 & 0.89 \\ \hline
Friends\_s03e05 & 0.78 & 0.88 \\ \hline
Friends\_s03e06 & 0.74 & 0.86 \\ \hline
\end{tabular}%
}
\end{table}


\section{Conclusions}
This work introduces a pilot dataset enlisting background characters in the TV show {\em Friends}. This dataset poses a significant advantage over existing datasets by the virtue of significantly higher number of face-tracks attributed to state-of-the-art face detectors. We introduced a strategy to construct robust audio-visual character profiles and imposed a constraint that speech and face from an active speaker instance match the same character profile. We used the character profiles to enhance active speaker localization and background character detection performance. 

We showed that the proposed background character detection strategy is limited by an upper bound on performance (Table~\ref{tab:bchar}). Further, it is indifferent to the source of faces, be it characters' faces or images of faces present in the video frames. One major next step to this work is attribute analysis for background characters in entertainment media. 
\newpage
\bibliographystyle{IEEEbib}
\bibliography{reference}
\flushend
\end{document}